\documentclass[11pt]{article}
\usepackage[margin=1in]{geometry}
\usepackage{amsmath,amssymb}
\usepackage{booktabs}
\usepackage{array}
\usepackage{graphicx}
\usepackage{xcolor}
\usepackage{hyperref}
\usepackage{cite}
\usepackage{tikz}
\usetikzlibrary{arrows.meta, positioning, fit}
\newcolumntype{L}[1]{>{\raggedright\arraybackslash}p{#1}}

\title{\textbf{ReCoLoRA}: Spectrum-Aware Recursive Consolidation for Continual LLM Fine-Tuning}
\author{\textbf{Wentao Lu}\thanks{luwentao1314@gmail.com}}

\begin{document}
\maketitle

\begin{abstract}
Parameter-efficient fine-tuning adapts a large language model to one task cheaply, but across a task sequence LoRA-style methods keep stacking low-rank updates on the same frozen weight, so each new task tends to overwrite the previous ones. We present \emph{ReCoLoRA} (Recursive Consolidation of Low-Rank Adapters), a spectrum-aware framework for continual fine-tuning: adapters are initialized from a randomized SVD of the pretrained weight, per-layer effective ranks are selected by an elbow criterion, and the principal subspace is adapted before residual capacity is opened. Before each new task, ReCoLoRA re-decomposes the \emph{current} effective weight, rather than the original one, into a frozen residual, a slowly updated principal component, and a fresh adapter (recursive consolidation), so every task starts from the model that has already absorbed its predecessors. On a six-task continual GLUE sequence over four 7--8B backbones, ReCoLoRA attains the best final average score on three of the four backbones against rank-swept LoRA, PiSSA, AdaLoRA, and DoRA baselines while training fewer parameters; an oracle-routed task-bank variant serves as an upper bound under full task isolation. Code: \url{https://github.com/bhqy666/ReCoLoRA}.
\end{abstract}

\section{Introduction}
Large language models (LLMs) are usually trained on static corpora, but deployed systems must adapt to changing tasks, domains, and user requirements. Re-training a full LLM whenever new data arrives is often infeasible, so downstream adaptation is commonly performed with PEFT. LoRA~\cite{hu2022lora} is a representative PEFT method: it freezes the pretrained weight matrix $W_0$ and learns a low-rank update
\begin{equation}
    W_{\mathrm{eff}} = W_0 + \Delta W, \qquad \Delta W = BA,
\end{equation}
where $B \in \mathbb{R}^{d_{out}\times r}$, $A \in \mathbb{R}^{r\times d_{in}}$, and $r \ll \min(d_{out}, d_{in})$. LoRA greatly reduces trainable parameters and is easy to deploy because the adapter can be merged into the base weight after training.

However, the usual single-task PEFT assumption is too weak for continual use. In continual learning, data or tasks arrive sequentially, and the learner must acquire new knowledge while retaining useful behavior from previous stages~\cite{wang2024comprehensivecl,ke2023continualnlp}. This creates the classical stability-plasticity dilemma: a model with high plasticity quickly adapts to a new task but may forget old tasks, while a highly stable model preserves old behavior but may underfit new tasks. Recent LLM studies show that catastrophic forgetting can also occur during continual instruction tuning and fine-tuning, including degradation in domain knowledge, reasoning, and reading-comprehension capabilities~\cite{luo2023empiricalforgetting,weng2024continualllm}. Thus, PEFT does not automatically solve forgetting; low-rank adapters can still over-specialize to the latest task.

Existing LoRA variants address related issues. AdaLoRA~\cite{zhang2023adalora} dynamically reallocates rank budgets during training. PiSSA~\cite{meng2024pissa} initializes adapters from principal singular vectors of pretrained weights, improving convergence over random LoRA initialization. DoRA~\cite{liu2024dora} separates weight magnitude and direction to improve adapter expressiveness. Orthogonal LoRA methods such as O-LoRA~\cite{wang2023olora} explicitly reduce interference by assigning new tasks to orthogonal low-rank subspaces. These methods motivate our central question: can we use the pretrained weight spectrum itself to decide which subspaces should change first, and which should be recovered later?

We propose ReCoLoRA (Recursive Consolidation of Low-Rank Adapters), a spectral and staged variant of LoRA for continual fine-tuning. ReCoLoRA decomposes each pretrained weight into a principal and a residual subspace, initializes the principal adapter with randomized SVD, selects an effective rank per layer with an elbow criterion, and trains in two stages: the first stage adapts the principal subspace, which carries the dominant pretrained structure, and the second stage gradually opens up residual capacity when the principal subspace alone cannot fit the task. The harder question is what to do across a sequence of tasks. A natural first attempt, which we tried, is to protect old knowledge inside a single shared adapter by freezing the ranks it has already used or by anchoring it in parameter space; our experiments show this is unreliable, because strong protection suppresses plasticity while weak protection lets the old behavior drift. The core of ReCoLoRA is therefore a different mechanism, recursive consolidation. Instead of forever adding new updates on top of the original frozen weight $W_0$, ReCoLoRA folds each finished task back into the model: before the next task it re-decomposes the current effective weight into a frozen residual, a slowly trainable elbow-compressed principal component, and a fresh fast adapter. Each task thus departs from the consolidated model, and the knowledge of earlier tasks is preserved in the slow principal subspace rather than left exposed to overwriting. The result is a single evolving model that can be deployed directly. To measure how much an explicit memory mechanism can recover, we also study ReCoLoRA-TaskBank, which trains one isolated spectrum-initialized ReCoLoRA branch per task, freezes finished branches, and routes each evaluation to its own branch with an oracle. TaskBank removes overwriting by construction; we read it as an upper bound rather than a deployable, task-agnostic solution.

Our contributions are:
\begin{itemize}
    \item We formulate ReCoLoRA, a spectrum-aware PEFT framework for continual fine-tuning that initializes adapters from a randomized SVD of the pretrained weight, selects an effective rank per layer with an elbow criterion, and adapts the principal subspace before the residual one.
    \item We introduce recursive consolidation as the continual mechanism of ReCoLoRA: each finished task is folded back into the model by re-decomposing the current effective weight into a frozen residual, a slowly trainable principal component, and a fresh adapter. This produces a single deployable model and, in our experiments, retains old tasks more reliably than freezing or anchoring ranks inside one shared adapter.
    \item We introduce ReCoLoRA-TaskBank, an oracle-routed task-incremental variant that isolates one spectrum-initialized branch per task. It removes overwriting by construction and serves as an upper bound on retention.
    \item On a six-task continual GLUE sequence over Qwen3-8B, Llama-3.1-8B-Instruct, Mistral-7B-v0.3, and InternLM2.5-7B-Chat, and against rank-swept baselines, ReCoLoRA achieves the best final average score on three of the four backbones with far fewer trainable parameters than the strongest rank-256 baseline, while ReCoLoRA-TaskBank attains $0.8957\pm0.0026$ final average score with $0.0000\pm0.0000$ average forgetting on Qwen3-8B across three seeds under oracle routing.
\end{itemize}

\section{Background and Related Work}
\subsection{Continual Learning Settings and Metrics}
Continual learning studies models that receive data in a sequence rather than as a fixed independent and identically distributed training set. General surveys distinguish several common settings~\cite{wang2024comprehensivecl}: instance-incremental learning changes the input stream within a task; domain-incremental learning changes input distributions while keeping label semantics similar; task-incremental learning introduces a sequence of tasks, often with task identity available at evaluation; class-incremental learning introduces new labels without task identity at test time; online continual learning restricts training to a single or small number of passes over streaming data. LLM continual learning is often organized by training stage: continual pretraining updates facts, domains, or languages; continual instruction tuning updates tasks, skills, and domains; continual alignment updates preferences or values~\cite{weng2024continualllm}.

This paper focuses on task-incremental continual fine-tuning. We sequentially train one model on a stream of downstream language understanding tasks. The model is evaluated after each stage on all tasks seen so far. This setting is close to continual instruction-style adaptation because each classification task is converted into a natural-language prompt and solved with a causal language model. It does not use replay data from previous tasks, and all baseline methods share the same task order and evaluation protocol.

Typical NLP/LLM continual learning benchmarks include GLUE and SuperGLUE-style task sequences~\cite{wang2019glue,wang2019superglue}, lifelong language learning streams such as the five-task LAMOL setting~\cite{sun2019lamol}, instruction-tuning collections such as T0/FLAN-style task mixtures~\cite{scialom2022continualt0}, and capability-retention suites that evaluate domain knowledge, reasoning, and reading comprehension after fine-tuning~\cite{luo2023empiricalforgetting}. For the first complete implementation we choose GLUE because it is public, lightweight enough for repeated 8B-model runs, and gives heterogeneous task formats: sentiment classification, paraphrase detection, entailment, question answering as entailment, duplicate-question detection, and multi-genre inference.

Let $a_{i,j}$ be the score on task $j$ after the model has trained through task $i$. The final average score after $T$ tasks is
\begin{equation}
    \mathrm{FinalAvg} = \frac{1}{T}\sum_{j=1}^{T} a_{T,j}.
\end{equation}
A standard forgetting measure compares the final score on each task with the best score observed earlier in the sequence~\cite{wang2024comprehensivecl}:
\begin{equation}
    F_j = \max_{i\in\{j,\ldots,T\}} a_{i,j} - a_{T,j},
    \qquad
    \mathrm{AvgForget} = \frac{1}{T}\sum_{j=1}^{T} F_j.
    \label{eq:avgforget}
\end{equation}
Note that $F_T=0$ by construction: the last task in the sequence is evaluated only once, immediately after its own training, so it cannot register forgetting regardless of method. AvgForget therefore depends not only on the method but also on which task occupies the last position.
We report FinalAvg and AvgForget, which capture the two dimensions most relevant to this paper: final utility and memory stability.

\subsection{PEFT and Low-Rank Adaptation}
LoRA~\cite{hu2022lora} freezes the backbone and learns low-rank updates. Subsequent work improves LoRA from several directions. AdaLoRA~\cite{zhang2023adalora} allocates rank adaptively by estimating component importance. DyLoRA~\cite{valipour2023dylora} supports multiple ranks from one trained adapter. DoRA~\cite{liu2024dora} decomposes pretrained weights into magnitude and direction. VeRA~\cite{kopiczko2024vera} reduces trainable parameters by sharing random projections. LoRA+~\cite{hayou2024loraplus} improves optimization by using different learning rates for LoRA factors.

\subsection{Continual Learning for NLP and LLMs}
Continual learning in NLP differs from vision because tasks may have different input-output formats, label semantics, and instruction styles~\cite{ke2023continualnlp}. LAMOL~\cite{sun2019lamol} is an early language lifelong learning method that trains a language model to solve tasks and generate pseudo-samples for replay. Continual-T0~\cite{scialom2022continualt0} studies continual instruction learning over many tasks and argues that instruction-tuned language models can retain broad skills under appropriate training. Progressive Prompts~\cite{razdaibiedina2023progressiveprompts} keeps the base model frozen and learns a new soft prompt for each task, concatenating prompts sequentially to reduce forgetting.

For LoRA-style continual learning, O-LoRA~\cite{wang2023olora} learns new tasks in orthogonal low-rank subspaces and fixes the old ones to limit interference, while C-LoRA~\cite{zhang2025clora} uses a routing matrix to manage parameter updates across tasks so that learned subspaces can be reused without maintaining many independent adapters. ReCoLoRA shares the view that subspace interference is the central problem, but addresses it differently: it starts from the pretrained weight spectrum and, after each task, recomputes a slow principal subspace from the current effective weight, so earlier knowledge is carried forward inside one evolving model. The TaskBank variant sits at the opposite end, close to adapter-isolation and routing methods: it allocates one spectrum-initialized branch per task and, in this paper, uses oracle routing as an upper bound before a learned, input-conditioned router is introduced.

\section{Method}
\begin{figure}[t]
\centering
\includegraphics[width=\linewidth]{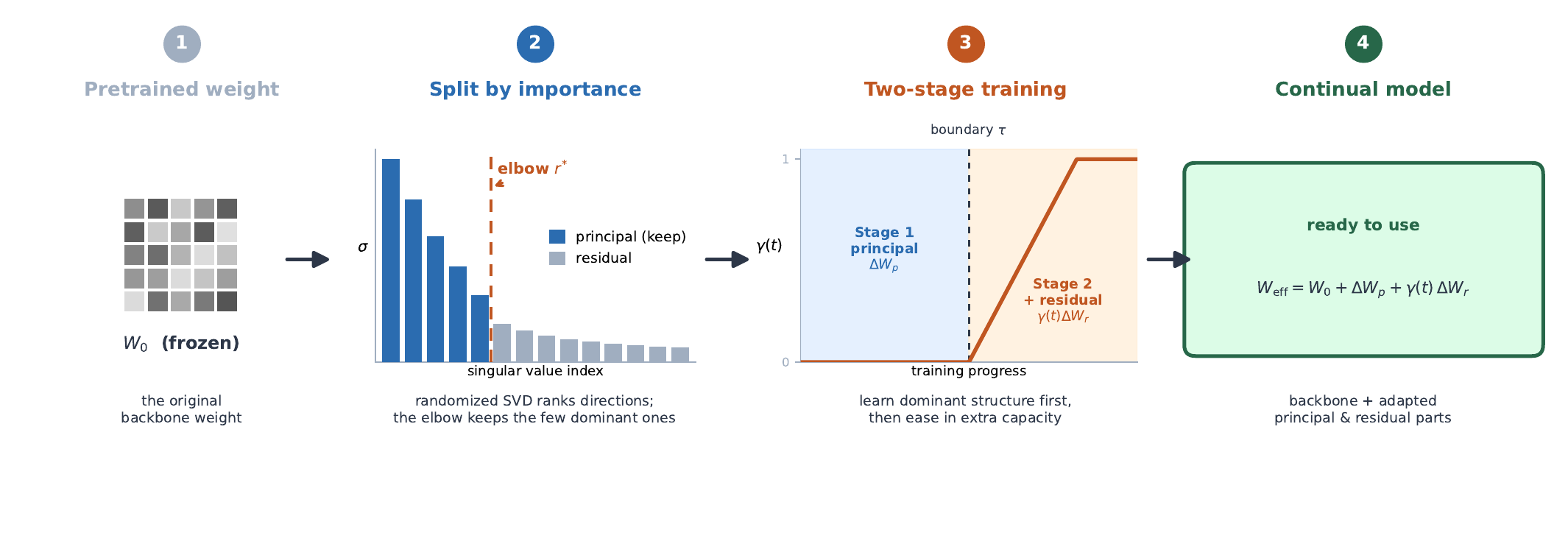}
\caption{ReCoLoRA pipeline. Each target pretrained weight is decomposed approximately with randomized SVD. An elbow heuristic chooses the effective rank and induces a rank mask. Training first adapts the principal subspace, then gradually recovers residual capacity through adapter parameters rather than unfreezing the full backbone.}
\label{fig:pipeline}
\end{figure}

ReCoLoRA modifies LoRA along three axes: spectral initialization, adaptive effective rank, and staged residual recovery. Figure~\ref{fig:pipeline} summarizes the pipeline.

The method is designed around two observations. First, pretrained transformer weights are not isotropic: their singular spectra often show a small number of dominant directions followed by a long tail. Second, catastrophic forgetting in sequential fine-tuning is caused not only by parameter count, but also by where the update is applied. Random low-rank factors can use their capacity in directions that fit the current task while interfering with previous behavior. ReCoLoRA therefore asks the optimizer to first use directions already important to the pretrained model, and only later use residual capacity.

\subsection{Spectral View of LoRA Updates}
Consider a pretrained matrix $W_0 \in \mathbb{R}^{d_{out}\times d_{in}}$ with singular value decomposition
\begin{equation}
    W_0 = U\Sigma V^\top = U_r\Sigma_rV_r^\top + U_{\perp}\Sigma_{\perp}V_{\perp}^\top.
\end{equation}
The first term contains the leading $r$ singular directions and is called the principal component. The remaining term is the residual component. Standard LoRA does not explicitly use this decomposition; its low-rank factors are randomly initialized and must discover useful directions through task gradients. PiSSA uses the principal singular directions for initialization~\cite{meng2024pissa}. ReCoLoRA follows the same spectral motivation but adds two continual-learning mechanisms: adaptive rank masking and residual recovery.

ReCoLoRA represents the effective weight as
\begin{equation}
    W_{\mathrm{eff}}(t) = W_0 + \Delta W_p(t) + \gamma(t)\Delta W_r(t),
\end{equation}
where $\Delta W_p$ is the principal adapter, $\Delta W_r$ is the residual adapter, and $\gamma(t)\in[0,1]$ controls when residual updates affect the forward pass. The backbone $W_0$ remains frozen. This is important: ReCoLoRA is not full fine-tuning of residual weights. It is still a PEFT method, but it separates the geometry of adapter updates into principal and residual components.

\subsection{Randomized SVD Initialization}
Direct full SVD is expensive for transformer weights. ReCoLoRA uses randomized SVD~\cite{halko2011finding}. For a target rank $r$ and oversampling parameter $p$, we sample a Gaussian matrix $\Omega \in \mathbb{R}^{d_{in}\times (r+p)}$ and form
\begin{equation}
    Y = W_0\Omega.
\end{equation}
After orthonormalizing $Y$ into $Q$, we project the weight into the smaller subspace,
\begin{equation}
    C = Q^\top W_0,
\end{equation}
compute an SVD of $C$, and reconstruct approximate leading singular vectors of $W_0$. Optional power iterations improve accuracy when the singular spectrum decays slowly. The cost is roughly $O(d_{out}d_{in}r)$, much smaller than full SVD when $r \ll \min(d_{out},d_{in})$.

Given approximate $U_r,\Sigma_r,V_r$, the principal adapter is initialized so that the initial low-rank update is aligned with $U_r\Sigma_rV_r^\top$. In practice, the adapter factors are scaled according to the LoRA convention so that the effective model remains numerically stable at the start of fine-tuning. This initialization gives the optimizer directions that already explain high-energy structure in the pretrained weight, reducing the need to discover them from random factors.

\subsection{Elbow-Based Rank Selection and Rank Masks}
A fixed rank applies the same capacity to every target matrix, even though transformer layers can have different spectral profiles. ReCoLoRA selects an effective rank from the singular values. Let $\sigma_1\geq\sigma_2\geq\cdots$ be the singular values of a target matrix. We search for an elbow point where the marginal spectral contribution drops sharply. A practical criterion combines a ratio test and an energy constraint:
\begin{equation}
    r^* = \arg\max_r \frac{\sigma_r}{\sigma_{r+1}}, \qquad
    \frac{\sum_{i=1}^{r^*}\sigma_i^2}{\sum_{i=1}^{r_{\max}+r_{res}}\sigma_i^2} \geq \rho,
    \label{eq:elbow}
\end{equation}
with clipping to a minimum and maximum rank. The energy ratio is normalized against the truncated spectral budget that randomized SVD actually computes ($r_{\max}+r_{res}$ singular values), not the full matrix spectrum: for large transformer weight matrices, the energy contained in the top $r_{\max}$ singular values is a vanishing fraction of the full-matrix energy, so a full-matrix normalization would make $\rho$ unreachable and collapse $r^*$ to $r_{\max}$ for every layer. The exact thresholds are implementation hyperparameters; in the current experiments the single-adapter setting uses a rank-16 budget with $\rho=0.8$, which yields per-layer principal ranks in $[8,16]$, while the TaskBank branches use a rank-32 allocated budget with a smaller active elbow rank per task.

\begin{figure}[t]
\centering
\includegraphics[width=\linewidth]{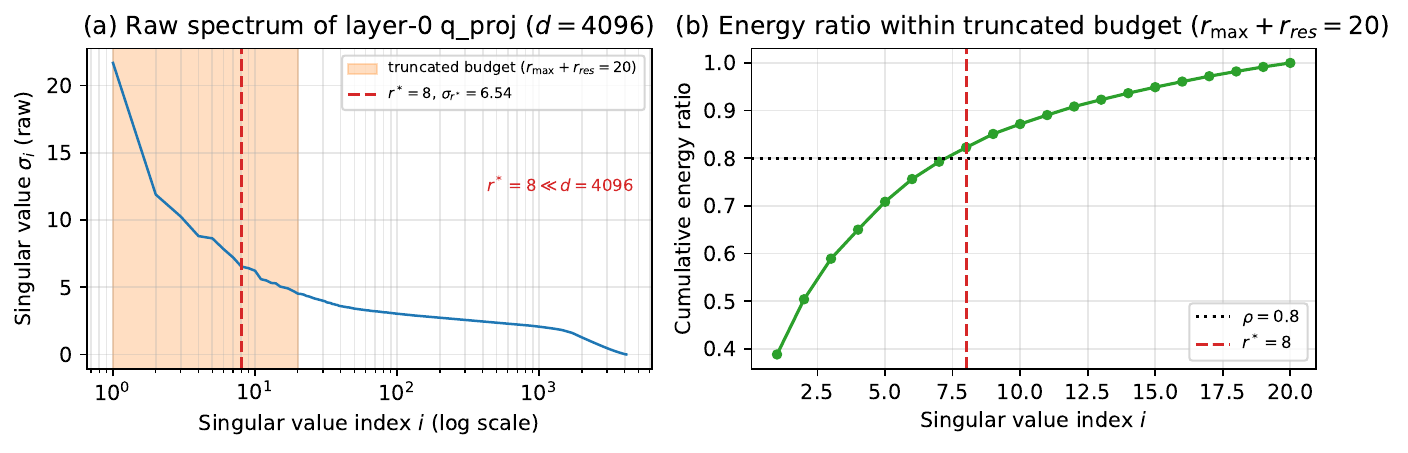}
\caption{Elbow/energy-based rank selection (Eq.~\ref{eq:elbow}) applied to a real Qwen3-8B weight matrix (\texttt{model.layers.0.self\_attn.q\_proj}, $d=4096$), using the same randomized SVD (truncated rank $r_{\max}+r_{res}=20$, oversample 5, seed 42) and selection hyperparameters ($\rho=0.8$, elbow ratio $1.5$, $r_{\max}=16$, $r_{\min}=2$) as the main experiments. (a) Raw (unnormalized) singular-value spectrum on a log-index/linear-value axis, in the classic ``knee'' style: $\sigma_i$ drops sharply over the first $\sim$10 indices and then flattens across the remaining $d-20\approx4076$ dimensions. The orange band marks the truncated computation budget $r_{\max}+r_{res}=20$ that randomized SVD actually computes, and the selected rank $r^*=8$ (with $\sigma_{r^*}=6.54$) falls well inside it ($r^*\ll d$), visually confirming that only a handful of directions carry most of the matrix's spectral mass. (b) Cumulative energy ratio within the truncated budget; $r^*$ is the smallest index at which this ratio reaches $\rho=0.8$. The rank mask keeps principal dimensions $1,\ldots,r^*$ active and treats the remaining allocated dimensions as residual capacity.}
\label{fig:elbow}
\end{figure}

The chosen rank is implemented as a rank mask over the allocated adapter dimensions. During the principal stage, only selected principal directions are active. During residual recovery, the mask and residual ratio can be updated to expose additional capacity. This is a lightweight mechanism for controlling plasticity without changing the backbone or allocating a new adapter for every task.

\subsection{Two-Stage Principal-to-Residual Training}
\label{sec:two-stage}
ReCoLoRA's training schedule is designed for the stability-plasticity trade-off. Stage 1 adapts only the principal adapter branch. Because these directions correspond to high-energy pretrained components, early updates remain close to the pretrained representation and are less likely to perturb low-energy directions that may encode fragile but useful behavior. Stage 2 activates residual recovery with a smaller residual learning rate or a scheduled residual ratio. This gives the model additional degrees of freedom after the principal branch has learned the broad task structure.

The schedule can be written as
\begin{equation}
    \gamma(t) =
    \begin{cases}
        0, & t < \tau,\\
        g(t), & t \geq \tau,
    \end{cases}
\end{equation}
where $\tau$ is the stage boundary and $g(t)$ increases the residual contribution. A simple step schedule is sufficient in the current implementation, but smoother schedules are compatible with the same parameterization. We use separate parameter groups so that residual parameters can receive a lower learning rate. This avoids a sudden high-variance update when residual capacity is introduced.

\subsection{Recursive Consolidation}
The mechanisms described so far still treat the original pretrained weight $W_0$ as a permanent reference. This is a useful PEFT constraint, but it is a poor model of continual adaptation: after a task is learned, the next task should start from the consolidated effective model rather than repeatedly attaching new updates to the same untouched $W_0$. ReCoLoRA therefore decomposes the effective weight before each task into a frozen residual, a slowly trainable principal component, and a fresh fast adapter:
\begin{equation}
    W_{\mathrm{eff}}^{(t)} = W_{\mathrm{res}}^{(t)} + W_{\mathrm{slow}}^{(t)} + \Delta W_{\mathrm{fast}}^{(t)}.
\end{equation}
For the first task, let $W_{\mathrm{eff}}^{(0)}=W_0$. Before task $t$, we compute an elbow rank $k_t$ from the spectrum of the previous effective weight and define
\begin{equation}
    W_{\mathrm{slow}}^{(t)} = \mathrm{topSVD}(W_{\mathrm{eff}}^{(t-1)}, k_t), \qquad
    W_{\mathrm{res}}^{(t)} = W_{\mathrm{eff}}^{(t-1)} - W_{\mathrm{slow}}^{(t)}.
\end{equation}
The slow component is stored as two low-rank factors and is updated with a much smaller effective learning rate or a delayed ramp, so dominant pretrained and previously consolidated directions change only gradually. The fast component is initialized near zero and receives most short-term task plasticity. Its rank is tied to the layer-specific elbow rank, for example $k_t+3$ or a small percentage increase clipped by the maximum-rank budget.

At the task boundary, the trained weight
\begin{equation}
    \widehat{W}^{(t)} = W_{\mathrm{res}}^{(t)} + W_{\mathrm{slow}}^{(t)} + \Delta W_{\mathrm{fast}}^{(t)}
\end{equation}
is consolidated by SVD again. The next task receives a new pair $(W_{\mathrm{res}}^{(t+1)}, W_{\mathrm{slow}}^{(t+1)})$ from $\widehat{W}^{(t)}$, and $\Delta W_{\mathrm{fast}}^{(t+1)}$ is reinitialized. Thus, low-value rank directions can be pruned, layers whose spectra require more capacity can grow, and the model no longer accumulates all tasks as unrelated deltas on the same original backbone. This variant remains parameter-efficient during each task, but it explicitly allows the consolidated effective weight to drift slowly over a task sequence.

\subsection{ReCoLoRA-TaskBank: Adapter Isolation with Oracle Routing}
Recursive consolidation lets one model evolve, but it still merges every task into a shared set of weights. When task identity is available at evaluation, a stronger guarantee is possible: give each task its own parameters so that later tasks cannot touch earlier ones. Approximating this inside a single shared adapter---by freezing or anchoring old ranks (Appendix~\ref{app:rankfreeze})---is unreliable, because hard freezing reduces plasticity, soft freezing lets old ranks drift, and parameter-space anchoring does not preserve old-task outputs. ReCoLoRA-TaskBank instead prevents overwriting by construction.

For a task sequence $\{D_1,\ldots,D_T\}$, ReCoLoRA-TaskBank trains a separate adapter branch $\phi_t$ for each task:
\begin{equation}
    f_t(x) = f_{\theta_0,\phi_t}(x), \qquad t\in\{1,\ldots,T\},
\end{equation}
where the backbone $\theta_0$ is frozen and each branch $\phi_t$ is initialized with the same ReCoLoRA procedure: randomized SVD, elbow-based effective rank, and principal-to-residual two-stage training. After task $t$ is trained, $\phi_t$ is frozen. Later tasks train only their own branches, so their updates cannot overwrite the parameters used by earlier tasks.

In this manuscript we evaluate the task-bank with oracle task routing: when evaluating task $j$, the model uses branch $\phi_j$. This matches the task-incremental continual-learning assumption where task identity is available at test time. It should be read as an upper-bound experiment rather than a complete task-agnostic deployment solution. The next step is to replace the oracle with a learned router or gated mixture that selects branches from the input prompt.

\begin{table}[t]
\centering
\begin{tabular}{L{0.95\textwidth}}
\toprule
\textbf{Algorithm 1: ReCoLoRA continual fine-tuning for a task sequence} \\
\midrule
\textbf{Input:} pretrained model $f_{\theta}$, target modules $\mathcal{M}$, task sequence $\mathcal{T}=\{D_1,\ldots,D_T\}$, maximum rank $r_{max}$, residual rank $r_{res}$, stage boundary $\tau$. \\
1. For each target matrix $W_0 \in \mathcal{M}$, compute approximate leading singular components using randomized SVD. \\
2. Select an effective rank $r^*$ using the elbow criterion and construct a rank mask over $r_{max}$ allocated dimensions. \\
3. Initialize the principal adapter from the leading singular components and initialize the residual adapter with low contribution. \\
4. For each task $D_t$ in the sequence, optionally recompute layer-wise elbow ranks, expand the active rank mask, and freeze ranks allocated to previous tasks. \\
5. Train Stage 1 with principal adapter updates active and residual contribution suppressed. \\
6. After the stage boundary, activate residual recovery with a smaller residual learning rate or scheduled residual ratio. \\
7. Consolidate the task-trained effective weight by SVD, recompute layer-wise elbow ranks, carry the compressed slow component into the next task, and reset the fast adapter (recursive consolidation, the default). \\
8. For ReCoLoRA-TaskBank, instantiate a separate ReCoLoRA branch for each task, freeze previous branches, and route evaluation to the corresponding branch. \\
9. After each task, evaluate all tasks seen so far and update the evaluation matrix $a_{i,j}$. \\
\textbf{Output:} final adapter parameters, task-by-stage evaluation matrix, FinalAvg and AvgForget. \\
\bottomrule
\end{tabular}
\caption{Decision-level ReCoLoRA training procedure. The backbone is frozen in the static and TaskBank variants; the default ReCoLoRA recursive consolidation instead freezes only the residual part of the previous consolidated effective weight while slowly updating the compressed principal component and training a fresh fast adapter.}
\label{tab:algorithm}
\end{table}

\section{Experimental Design}
\subsection{Continual GLUE Protocol}
We evaluate causal language models using a sequential GLUE task stream~\cite{wang2019glue}. The main experiments use Qwen3-8B, and the same protocol is repeated on Llama-3.1-8B-Instruct as a backbone validation. A focused Llama-3-8B study (Appendix~\ref{app:llama}) tests the task-wise rank-freezing mechanism after we observe that simple rank expansion alone does not reliably reduce forgetting. Finally, we evaluate ReCoLoRA against LoRA, PiSSA, AdaLoRA, and DoRA on Mistral-7B-v0.3 and InternLM2.5-7B-Chat to test whether recursive consolidation transfers beyond the Qwen/Llama experiments. The task order is
\begin{equation}
    \mathrm{SST\mbox{-}2} \rightarrow \mathrm{MRPC} \rightarrow \mathrm{QNLI} \rightarrow \mathrm{RTE} \rightarrow \mathrm{QQP} \rightarrow \mathrm{MNLI}.
\end{equation}
Each classification task is converted into a prompt with verbalized candidate labels. During training, the model is optimized to generate the correct label token sequence. During evaluation, each candidate label is scored by sequence loss and the lowest-loss label is selected. After training on task $i$, the model is evaluated on all tasks $1,\ldots,i$, giving a lower-triangular evaluation matrix.

The resulting matrix has the form shown in Table~\ref{tab:eval-matrix}. Entries above the diagonal are intentionally blank because future tasks are not evaluated before they are introduced in the current protocol. The diagonal measures immediate plasticity on the newly learned task, while changes down each column measure forgetting or backward transfer.

\begin{table}[t]
\centering
\begin{tabular}{lcccccc}
\toprule
After training & SST-2 & MRPC & QNLI & RTE & QQP & MNLI \\
\midrule
SST-2 & $a_{1,1}$ & -- & -- & -- & -- & -- \\
MRPC & $a_{2,1}$ & $a_{2,2}$ & -- & -- & -- & -- \\
QNLI & $a_{3,1}$ & $a_{3,2}$ & $a_{3,3}$ & -- & -- & -- \\
RTE & $a_{4,1}$ & $a_{4,2}$ & $a_{4,3}$ & $a_{4,4}$ & -- & -- \\
QQP & $a_{5,1}$ & $a_{5,2}$ & $a_{5,3}$ & $a_{5,4}$ & $a_{5,5}$ & -- \\
MNLI & $a_{6,1}$ & $a_{6,2}$ & $a_{6,3}$ & $a_{6,4}$ & $a_{6,5}$ & $a_{6,6}$ \\
\bottomrule
\end{tabular}
\caption{Task-by-stage evaluation matrix used in the continual GLUE protocol. FinalAvg uses the last row; AvgForget compares each column's final value with its best previous value.}
\label{tab:eval-matrix}
\end{table}

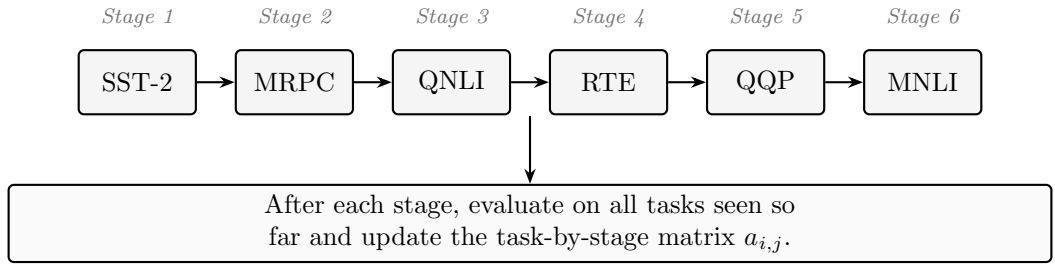
\begin{figure}[t]
\centering
\begin{tikzpicture}[
  font=\small,
  task/.style={draw, rounded corners=2pt, fill=black!4, minimum width=1.55cm,
               minimum height=0.85cm, align=center, thick},
  stage/.style={font=\scriptsize\itshape, text=black!55},
  arr/.style={-{Stealth[length=2mm]}, thick},
  node distance=5mm
]
  \node[task] (t1) {SST-2};
  \node[task, right=of t1] (t2) {MRPC};
  \node[task, right=of t2] (t3) {QNLI};
  \node[task, right=of t3] (t4) {RTE};
  \node[task, right=of t4] (t5) {QQP};
  \node[task, right=of t5] (t6) {MNLI};

  \draw[arr] (t1) -- (t2);
  \draw[arr] (t2) -- (t3);
  \draw[arr] (t3) -- (t4);
  \draw[arr] (t4) -- (t5);
  \draw[arr] (t5) -- (t6);

  \foreach \i/\n in {1/t1, 2/t2, 3/t3, 4/t4, 5/t5, 6/t6}{
    \node[stage, above=1.5mm of \n] {Stage \i};
  }

  \node[fit=(t1)(t6), inner sep=0pt] (chain) {};
  \node[draw, rounded corners=2pt, fill=black!2, align=center, thick,
        text width=0.82\textwidth, below=9mm of chain] (evalbox)
    {After each stage, evaluate on all tasks seen so far and update the task-by-stage matrix $a_{i,j}$.};

  \draw[arr] (chain.south) -- (evalbox.north);
\end{tikzpicture}
\caption{Continual fine-tuning protocol. The model sees one downstream task at a time. After every stage, all previously seen tasks are evaluated, which allows us to compute final average score, average forgetting, and transfer-related metrics.}
\label{fig:continual_protocol}
\end{figure}

\subsection{Baselines}
We compare the following PEFT methods:
\begin{itemize}
    \item \textbf{LoRA}: standard fixed-rank low-rank adaptation with random initialization.
    \item \textbf{PiSSA}: principal singular vector initialization for LoRA adapters~\cite{meng2024pissa}.
    \item \textbf{AdaLoRA}: adaptive rank allocation during training~\cite{zhang2023adalora}.
    \item \textbf{DoRA}: weight-decomposed LoRA with magnitude-direction separation~\cite{liu2024dora}.
    \item \textbf{ReCoLoRA}: the full method---randomized SVD initialization, elbow-based rank control, staged principal-to-residual training, and recursive consolidation, which re-decomposes the effective weight at each task boundary into a frozen residual, an elbow-compressed slow principal subspace, and a fresh fast adapter.
    \item \textbf{Static ReCoLoRA}: ReCoLoRA without recursive consolidation (a single shared adapter whose updates always sit on the original $W_0$); used as an ablation (Appendix~\ref{app:ablations}).
    \item \textbf{ReCoLoRA-TaskBank}: an oracle-routed task-incremental variant that trains an isolated ReCoLoRA branch for each task and freezes old branches.
\end{itemize}
All methods use the same task order and random seeds. For Qwen, Llama, and Mistral backbones, adapters target the query and value projection modules; for InternLM2.5-7B-Chat, the adapters target the fused attention projection and output projection modules used by the local model implementation. Each task is trained for 200 optimizer steps, with maximum sequence length 256 and at most 1000 evaluation samples per task. We report mean and standard deviation over seeds 42, 43, and 44. TaskBank uses oracle task identity at evaluation, so it is reported as a task-incremental upper bound rather than as a task-agnostic router.

\section{Results}
\subsection{Main Continual Fine-Tuning Results}
Table~\ref{tab:qwen3-main} reports the Qwen3-8B comparison of single-adapter PEFT methods, O-LoRA (a representative continual-learning LoRA baseline), and the oracle-routed TaskBank upper bound over three seeds. Among the single-adapter methods, static ReCoLoRA obtains the highest final average score and reduces forgetting relative to LoRA, PiSSA, DoRA, and O-LoRA, confirming that spectral initialization with staged training improves single-adapter quality. O-LoRA, which trains one frozen LoRA branch per task with inter-task orthogonality regularization, achieves intermediate performance: it outperforms LoRA on forgetting (0.0257 vs 0.0414) but falls short of ReCoLoRA on final average score (0.8722 vs 0.8821) using substantially more stored parameters ($\sim$23M vs 7.67M for six tasks). ReCoLoRA-TaskBank then eliminates measured forgetting under oracle routing, isolating branch isolation as the component that structurally prevents overwriting. The deployable ReCoLoRA with recursive consolidation is evaluated across backbones in Table~\ref{tab:recursive-backbones}.

\begin{table}[t]
\centering
\begin{tabular}{lccc}
\toprule
Method & Final Avg. $\uparrow$ & Avg. Forgetting $\downarrow$ & Trainable Params \\
\midrule
AdaLoRA & 0.8597 $\pm$ 0.0017 & 0.0133 $\pm$ 0.0010 & 11,503,368 \\
DoRA & 0.8551 $\pm$ 0.0055 & 0.0487 $\pm$ 0.0058 & 7,852,032 \\
Static ReCoLoRA & 0.8821 $\pm$ 0.0058 & 0.0240 $\pm$ 0.0054 & 6,850,688$^{*}$ \\
ReCoLoRA-TaskBank (oracle) & \textbf{0.8957 $\pm$ 0.0026} & \textbf{0.0000 $\pm$ 0.0000} & 15,336,576/branch \\
LoRA & 0.8634 $\pm$ 0.0071 & 0.0414 $\pm$ 0.0082 & 7,667,712 \\
O-LoRA & 0.8722 $\pm$ 0.0064 & 0.0257 $\pm$ 0.0060 & $\sim$23,003,136 \\
PiSSA & 0.8621 $\pm$ 0.0161 & 0.0418 $\pm$ 0.0163 & 7,667,712 \\
\bottomrule
\end{tabular}
\caption{Qwen3-8B PEFT continual fine-tuning results on the GLUE task sequence SST-2 $\rightarrow$ MRPC $\rightarrow$ QNLI $\rightarrow$ RTE $\rightarrow$ QQP $\rightarrow$ MNLI. Results are mean $\pm$ standard deviation over seeds 42, 43, and 44. ReCoLoRA-TaskBank uses oracle task routing and should be interpreted as a task-incremental upper bound; it stores one branch per task. O-LoRA stores one low-rank adapter per task at rank 8, so memory scales with task count; the value shown is approximate for six tasks. $^{*}$Static ReCoLoRA's trainable parameter count varies slightly by seed (6,848,640--6,851,712) because the elbow/energy rank selector ($\rho \geq 0.8$ of the truncated-spectrum energy, $r_{\max}=16$) assigns a different principal rank $r\in[8,16]$ to each projection matrix; the value reported is the across-seed average.}
\label{tab:qwen3-main}
\end{table}

\subsection{Comparison with Continual Learning Baselines}
Table~\ref{tab:mistral-single} extends the single-adapter comparison to Mistral-7B-v0.3 using the same continual-GLUE protocol. This enables a backbone-level robustness check for static ReCoLoRA against O-LoRA, a method designed explicitly for continual learning. On Mistral, the two methods show different relative standings: O-LoRA achieves higher final average score and lower forgetting (0.8255 / 0.0675 vs 0.8031 / 0.0912), while both outperform standard LoRA (0.8044 / 0.0959). This result illustrates why ReCoLoRA research shifted toward recursive consolidation: the static two-stage schedule, while effective on Qwen3-8B, does not generalize robustly across backbones. The recursive variant (evaluated in Table~\ref{tab:recursive-backbones}) is designed to mitigate this sensitivity by consolidating and recomputing the task-specific schedule after each task.

\begin{table}[t]
\centering
\begin{tabular}{lccc}
\toprule
Method & Final Avg. $\uparrow$ & Avg. Forgetting $\downarrow$ & Trainable Params \\
\midrule
LoRA & 0.8044 $\pm$ 0.0155 & 0.0959 $\pm$ 0.0182 & 10,944,512 \\
ReCoLoRA (static, single-adapter) & 0.8031 $\pm$ 0.0139 & 0.0912 $\pm$ 0.0190 & 10,944,512$^{\dagger}$ \\
O-LoRA & 0.8255 $\pm$ 0.0067 & 0.0675 $\pm$ 0.0092 & $\sim$13,123,584 \\
\bottomrule
\end{tabular}
\caption{Mistral-7B-v0.3 single-adapter PEFT comparison on the six-task continual GLUE sequence. Results are mean $\pm$ standard deviation over seeds 42, 43, and 44. ReCoLoRA uses the same configuration as the Qwen3-8B main results (elbow rank selection with $\rho=0.8$, $r_{\max}=16$, two-stage training). O-LoRA trains one rank-8 adapter per task with orthogonality regularization. The higher forgetting for static ReCoLoRA on Mistral (0.0912) compared to Qwen3-8B (0.0240) suggests that the static schedule is backbone-sensitive; recursive consolidation (Table~\ref{tab:recursive-backbones}) is designed to address this. $^{\dagger}$ReCoLoRA parameter count varies by seed ($\sim$10.4--11.2M); value shown is approximate.}
\label{tab:mistral-single}
\end{table}

\subsection{Recursive Consolidation on Multiple Backbones}
Table~\ref{tab:recursive-backbones} evaluates ReCoLoRA against rank-swept baselines on four 7B/8B-class backbones. To address the possibility that ReCoLoRA benefits only from using a larger adaptive rank, each baseline is swept over ranks 64, 128, and 256, and the table reports the best final-average configuration for each method. ReCoLoRA is adaptive: each layer chooses an elbow-compressed slow rank with minimum rank 8 and maximum rank 256, then adds a fresh fast update for the next task.

The result is favorable but not universal. Under the seed-42 comparison reported in the table, ReCoLoRA obtains the best final average score on Qwen3-8B, Mistral-7B-v0.3, and InternLM2.5-7B-Chat, with its largest apparent gain on Mistral (0.0654 final-average points over the best baseline) while using fewer trainable parameters than all rank-64 LoRA-family baselines. On Qwen3-8B and InternLM2.5-7B-Chat the gains over the best LoRA-family baseline are smaller but still positive, and Llama-3.1-8B-Instruct appears to be a negative case, with LoRA at rank 64 ahead on final average score.

Because the baseline sweeps in Table~\ref{tab:recursive-backbones} use a single seed, we additionally reran ReCoLoRA (only) with seeds 43 and 44 on all four backbones to check how robust these single-seed comparisons are. The three-seed means (population standard deviation) for final average score are: Qwen3-8B $0.8863\pm0.0004$, Mistral-7B-v0.3 $0.8324\pm0.0303$, InternLM2.5-7B-Chat $0.8746\pm0.0049$, and Llama-3.1-8B-Instruct $0.8465\pm0.0104$. Qwen3-8B is essentially seed-independent, and InternLM2.5-7B-Chat's gain over LoRA narrows from 0.0055 to roughly parity but remains consistent with the seed-42 ranking. Mistral-7B-v0.3 is the most seed-sensitive backbone: the seed-42 gain of 0.0654 over the best baseline (LoRA at rank 64, 0.7979) shrinks to about 0.0345 on average across seeds, and one seed (0.7913) falls slightly below that baseline on final average score; ReCoLoRA's forgetting nevertheless remains the lowest in the table on average ($0.0590\pm0.0399$, versus $0.1007$--$0.1109$ for the Mistral baselines), so we present the Mistral result as favorable but seed-sensitive rather than a robust large gain. Conversely, Llama-3.1-8B-Instruct is less of a negative case than the single seed suggests: LoRA at rank 64 remains nominally ahead on final average score (0.8522 versus a three-seed mean of $0.8465\pm0.0104$), but two of the three seeds (0.8522 and 0.8554) match or exceed LoRA, and ReCoLoRA's three-seed average forgetting ($0.0453\pm0.0125$) is lower than LoRA's (0.0551). We therefore interpret ReCoLoRA as a promising single-evolving-model mechanism whose advantage is robust on two of the four backbones (Qwen3-8B, InternLM2.5-7B-Chat), favorable but seed-sensitive on a third (Mistral-7B-v0.3), and close to a tie rather than a clear loss on the fourth (Llama-3.1-8B-Instruct), rather than a universally dominant adapter.

\begin{table}[t]
\centering
\resizebox{\textwidth}{!}{
\begin{tabular}{llrcc}
\toprule
Backbone & Method & Trainable & Final Avg. $\uparrow$ & Avg. Forgetting $\downarrow$ \\
\midrule
Qwen3-8B & LoRA (r=64) & 30.7M & 0.8804 & 0.0290 \\
Qwen3-8B & PiSSA (r=64) & 30.7M & 0.8391 & 0.0673 \\
Qwen3-8B & AdaLoRA (r=64) & 46.0M & 0.8577 & 0.0205 \\
Qwen3-8B & DoRA (r=128) & 61.5M & 0.8779 & 0.0276 \\
Qwen3-8B & ReCoLoRA & 34.9M & \textbf{0.8866} & \textbf{0.0135} \\
\midrule
Mistral-7B-v0.3 & LoRA (r=64) & 27.3M & 0.7979 & 0.1053 \\
Mistral-7B-v0.3 & PiSSA (r=256) & 109.1M & 0.5237 & 0.0225 \\
Mistral-7B-v0.3 & AdaLoRA (r=256) & 163.6M & 0.7862 & 0.1007 \\
Mistral-7B-v0.3 & DoRA (r=64) & 27.4M & 0.7920 & 0.1109 \\
Mistral-7B-v0.3 & ReCoLoRA & 23.7M & \textbf{0.8634} & \textbf{0.0163} \\
\midrule
InternLM2.5-7B-Chat & LoRA (r=64) & 37.7M & 0.8747 & 0.0242 \\
InternLM2.5-7B-Chat & PiSSA (r=64) & 37.7M & 0.5229 & 0.0207 \\
InternLM2.5-7B-Chat & AdaLoRA (r=64) & 56.6M & 0.8594 & \textbf{0.0101} \\
InternLM2.5-7B-Chat & DoRA (r=128) & 75.8M & 0.8616 & 0.0520 \\
InternLM2.5-7B-Chat & ReCoLoRA & 35.3M & \textbf{0.8802} & 0.0177 \\
\midrule
Llama-3.1-8B-Instruct & LoRA (r=64) & 27.3M & \textbf{0.8522} & 0.0551 \\
Llama-3.1-8B-Instruct & PiSSA (r=64) & 27.3M & 0.7985 & 0.1000 \\
Llama-3.1-8B-Instruct & AdaLoRA (r=64) & 40.9M & 0.8216 & \textbf{0.0336} \\
Llama-3.1-8B-Instruct & DoRA (r=256) & 109.2M & 0.8177 & 0.0923 \\
Llama-3.1-8B-Instruct & ReCoLoRA & 25.4M & 0.8320 & 0.0621 \\
\bottomrule
\end{tabular}}
\caption{ReCoLoRA and rank-swept PEFT baselines on the six-task continual GLUE sequence. For each baseline method and backbone, ranks 64, 128, and 256 are swept and the row reports the rank with the highest final average score. Results in this table use seed 42 for all methods; a three-seed robustness check for ReCoLoRA is discussed in the text below the table. AdaLoRA at rank 256 exceeded our GPU memory budget for Qwen3-8B and InternLM2.5-7B-Chat, so the best completed rank is reported for those backbones instead. ReCoLoRA uses elbow-only consolidation with minimum rank 8 and maximum rank 256.}
\label{tab:recursive-backbones}
\end{table}

\subsection{Interpreting the Stability-Plasticity Trade-Off}
The Qwen3-8B table shows why continual PEFT should not be judged by a single metric. Among single-adapter methods, static ReCoLoRA gives the strongest final adaptation result while keeping forgetting low. Compared with standard LoRA and PiSSA on Qwen3-8B, static ReCoLoRA improves both final score and forgetting, which supports the claim that principal initialization should be paired with a staged residual strategy rather than used only as a static initialization trick. However, the TaskBank result changes the main interpretation: explicit adapter isolation is much more effective than trying to preserve all tasks inside one shared adapter. With oracle routing, ReCoLoRA-TaskBank obtains $0.8957\pm0.0026$ final average score and zero measured forgetting across all three seeds.

The recursive results sharpen the interpretation. Keeping all learning as a delta on the original $W_0$ is not the only PEFT-compatible design. By recomputing the slow principal component from the previous task's effective weight, ReCoLoRA gives the next task a consolidated starting point while still pruning low-value rank directions. The rank sweep is important: simply increasing LoRA-family ranks to 128 or 256 does not consistently improve continual learning, and in several cases it worsens final average score or forgetting. ReCoLoRA is strongest on Qwen3-8B, Mistral-7B-v0.3, and InternLM2.5-7B-Chat, but Llama-3.1-8B-Instruct remains a boundary case where standard LoRA is better. The strongest current evidence therefore points to spectrum-aware adaptation plus an explicit memory mechanism, either branch isolation for task-incremental upper-bound evaluation or recursive consolidation for a single evolving model, while also showing that the recursive schedule is backbone-sensitive.

\section{Discussion}
\subsection{Relation to Replay, Prompt, and Orthogonal-Subspace Methods}
Replay-based methods such as LAMOL reduce forgetting by regenerating or storing approximations of old data~\cite{sun2019lamol}, and prompt-based methods such as Progressive Prompts freeze the backbone and add per-task prompt parameters~\cite{razdaibiedina2023progressiveprompts}. Closer to ReCoLoRA, O-LoRA reserves orthogonal low-rank subspaces for different tasks~\cite{wang2023olora} and C-LoRA routes continual parameter updates~\cite{zhang2025clora}. ReCoLoRA differs in what it controls: the spectral organization of each pretrained weight. Its recursive consolidation keeps a single evolving model and folds each finished task into a slow principal subspace, so old knowledge is carried forward rather than protected by an external buffer or router. ReCoLoRA-TaskBank sits at the isolation end of the same spectrum, giving each task its own spectrum-initialized branch, and bounds how much retention is achievable. The two views separate the roles our experiments find necessary: spectral initialization improves how well each task is learned, while an explicit memory structure--a consolidated slow weight or an isolated branch--is what prevents overwriting.

\subsection{Limitations}
\label{sec:limitations}
Our scope is intentionally bounded. First, ReCoLoRA-TaskBank relies on oracle task routing and stores one branch per task: it is a valid task-incremental upper bound, but task-agnostic deployment and long task streams will require a learned router and some form of branch compression or merging. Second, the deployable recursive model is backbone-sensitive. Its advantage is robust on Qwen3-8B and InternLM2.5-7B-Chat, favorable but seed-sensitive on Mistral-7B-v0.3, and close to a tie on Llama-3.1-8B-Instruct, where rank-64 LoRA remains nominally ahead; we therefore do not claim a universally dominant adapter, and the SVD rank thresholds, residual learning rates, and slow-update schedule may need to be tuned per model family. Third, the task sequence is GLUE-style classification evaluated on a single task order; because AvgForget assigns zero forgetting to whichever task is evaluated last (Eq.~\ref{eq:avgforget}), longer instruction-following and reasoning streams, together with more task permutations and order-robust metrics, are needed before drawing order-independent conclusions.

\section{Conclusion}
We presented ReCoLoRA, a spectrum-aware PEFT framework for continual LLM fine-tuning. ReCoLoRA initializes adapters from a randomized SVD of the pretrained weight, selects an effective rank per layer with an elbow criterion, and adapts the principal subspace before the residual one; for continual use it consolidates each task recursively, re-decomposing the current effective weight into a frozen residual, a slow principal component, and a fresh adapter so that the model evolves as a single deployable network instead of a growing stack of updates on the original weight. We found that the simpler alternative, protecting old ranks inside one shared adapter through freezing or parameter-space anchoring, only trades plasticity for stability and does not reliably stop forgetting. On a six-task continual GLUE sequence over four 7--8B backbones, ReCoLoRA attains the best final average score on Qwen3-8B, Mistral-7B-v0.3, and InternLM2.5-7B-Chat with far fewer trainable parameters than the strongest rank-256 baseline, with Llama-3.1-8B-Instruct a close negative case. As an isolation upper bound under oracle routing, ReCoLoRA-TaskBank reaches $0.8957\pm0.0026$ final average score with $0.0000\pm0.0000$ average forgetting on Qwen3-8B, the ceiling that the single evolving model approaches. The picture that emerges is that spectral initialization improves how well each task is learned, while an explicit memory mechanism--recursive consolidation in one model, or branch isolation as a ceiling--is what keeps earlier tasks from being erased. Future work will add a learned router to remove the oracle assumption, branch compression for long streams, and evaluation on more task orders and longer instruction-following sequences.

\appendix

\noindent\emph{Note.} Throughout the appendix, ``ReCoLoRA'' without qualification denotes the static single-adapter variant (\emph{Static ReCoLoRA} in the main text); the deployable recursive method is the main-text ReCoLoRA.

\section{Single-Adapter Rank Freezing}\label{app:rankfreeze}
The original staged schedule controls when residual capacity is introduced, but it does not by itself prevent later tasks from modifying principal ranks that were useful for earlier tasks. To better match the continual-learning objective, we add an optional task-wise rank-freezing rule. At the beginning of task $t$, each target layer recomputes an elbow rank on its current effective weight and expands the active rank mask when additional capacity is needed. Let $m_{\ell,t}\in\{0,1\}^{r_{max}}$ denote the active rank mask of layer $\ell$ after task $t$ is introduced, and let $g_{\ell,t}$ denote the gradient mask. For $t>1$, old active ranks remain in the forward pass but are excluded from gradient updates:
\begin{equation}
    m_{\ell,t} \supseteq m_{\ell,t-1}, \qquad
    g_{\ell,t,k} =
    \begin{cases}
        1, & m_{\ell,t,k}=1 \ \mathrm{and}\ m_{\ell,t-1,k}=0,\\
        0, & \mathrm{otherwise}.
    \end{cases}
\end{equation}
In implementation, this is applied by masking gradients of the corresponding rows of $A$ and columns of $B$. The forward computation still uses all active ranks, so previous task subspaces remain available at inference time. New tasks are therefore encouraged to use newly opened dimensions instead of overwriting old ones. This mechanism is related in spirit to orthogonal-subspace continual learning, but it does not require a separate adapter per task; it grows a shared rank budget inside the same adapter.

\paragraph{Forward-preserving rank growth.}
Naively expanding the active rank mask can change the model function even before optimization if the frozen base weight already contains the SVD directions that are later reactivated through the adapter. To avoid this drift, the forward-preserving variant allocates a maximum SVD rank budget at initialization, subtracts the full allocated principal component from the frozen base, and keeps all allocated SVD directions present in the forward pass. The active-rank schedule then controls only which dimensions receive gradients. In a local sanity check, naive growth from rank 4 to rank 8 changed layer outputs by $8.8\times 10^{-1}$, while forward-preserving growth kept the output difference below $2\times 10^{-7}$. This distinction is important for continual learning because rank expansion should add trainable capacity without silently perturbing previously learned behavior.

\section{Single-Adapter Validation on Llama Backbones}\label{app:llama}
Table~\ref{tab:llama-main} reports the Llama-3.1-8B-Instruct validation under the same task sequence, seeds, and training budget. The result is intentionally reported as a robustness check rather than as a confirmation-only experiment. On this backbone, LoRA obtains the best final average score, with DoRA close behind. The original ReCoLoRA configuration improves over PiSSA and AdaLoRA in final average score but has higher forgetting. The forward-preserving frozen-rank variant reverses this behavior: it obtains the lowest forgetting in the table, but sacrifices final average score. This cross-backbone result narrows the claim: ReCoLoRA-style subspace preservation can reduce forgetting, but the current implementation is not yet a universally strongest PEFT method across all 8B backbones.

\begin{table}[t]
\centering
\begin{tabular}{lccc}
\toprule
Method & Final Avg. $\uparrow$ & Avg. Forgetting $\downarrow$ & Trainable Params \\
\midrule
AdaLoRA & 0.7932 $\pm$ 0.0096 & 0.0525 $\pm$ 0.0093 & 10,225,216 \\
DoRA & 0.8407 $\pm$ 0.0123 & 0.0577 $\pm$ 0.0137 & 6,979,584 \\
ReCoLoRA & 0.8310 $\pm$ 0.0109 & 0.0719 $\pm$ 0.0167 & 6,162,091$^{*}$ \\
ReCoLoRA-FP freeze & 0.8214 $\pm$ 0.0050 & \textbf{0.0378 $\pm$ 0.0066} & 10,224,640 \\
LoRA & \textbf{0.8427 $\pm$ 0.0162} & 0.0529 $\pm$ 0.0181 & 6,815,744 \\
PiSSA & 0.8117 $\pm$ 0.0103 & 0.0959 $\pm$ 0.0153 & 6,815,744 \\
\bottomrule
\end{tabular}
\caption{Llama-3.1-8B-Instruct PEFT continual fine-tuning results on the same GLUE task sequence. Results are mean $\pm$ standard deviation over seeds 42, 43, and 44. ReCoLoRA-FP freeze denotes the forward-preserving task-wise rank-freezing variant. It lowers forgetting but also reduces final average performance, exposing the stability-plasticity trade-off. $^{*}$ReCoLoRA's trainable parameter count varies slightly by seed (6,151,168--6,175,744) because the elbow/energy rank selector ($\rho \geq 0.8$ of the truncated-spectrum energy, $r_{\max}=16$) assigns a different principal rank $r\in[8,16]$ to each projection matrix; the value reported is the across-seed average.}
\label{tab:llama-main}
\end{table}

\begin{table}[t]
\centering
\begin{tabular}{lccc}
\toprule
Variant & Final Avg. $\uparrow$ & Avg. Forgetting $\downarrow$ & Trainable Params \\
\midrule
Task-wise elbow & 0.8095 $\pm$ 0.0041 & 0.0631 $\pm$ 0.0036 & 10,224,640 \\
Layer-wise bonus & 0.7984 $\pm$ 0.0094 & 0.0900 $\pm$ 0.0114 & 13,632,512 \\
Frozen old ranks & 0.8178 $\pm$ 0.0154 & \textbf{0.0283 $\pm$ 0.0187} & 13,632,512 \\
Forward-preserving frozen ranks & \textbf{0.8193 $\pm$ 0.0113} & 0.0340 $\pm$ 0.0141 & 13,632,512 \\
\bottomrule
\end{tabular}
\caption{Llama-3-8B ReCoLoRA rank-allocation study on the continual GLUE sequence. All variants use randomized SVD initialization and task-wise elbow recomputation. Task-wise elbow expands the active rank mask across tasks. Layer-wise bonus adds projection-type and depth-dependent rank bonuses. Frozen old ranks keeps previously active ranks in the forward pass but masks their gradients. Forward-preserving frozen ranks additionally keeps allocated inactive SVD directions present in the forward computation, so rank growth changes trainability rather than the initial function. Results are mean $\pm$ standard deviation over seeds 42, 43, and 44.}
\label{tab:llama-rank-freezing}
\end{table}

The Llama-3.1-8B-Instruct validation adds an important caveat. LoRA and DoRA remain stronger in final average score, but the forward-preserving frozen-rank variant has the lowest average forgetting. Compared with the earlier ReCoLoRA implementation, it reduces forgetting from 0.0719 to 0.0378, while reducing final average score from 0.8310 to 0.8214. This suggests that the spectral schedule is sensitive to backbone spectra, tokenizer-label interactions, or optimization hyperparameters, and it exposes a clear stability-plasticity trade-off.

Table~\ref{tab:llama-rank-freezing} clarifies the Llama-side failure mode. Simply recomputing elbow ranks per task gives moderate final performance but still leaves substantial forgetting. Adding more rank capacity by hand through layer-wise and depth-wise bonuses is not sufficient; it actually increases forgetting in this run. Freezing old ranks changes the behavior: the final average score improves from 0.8095 to 0.8178 relative to task-wise elbow, and average forgetting drops from 0.0631 to 0.0283. The forward-preserving version obtains a similar result, with slightly higher final average score (0.8193) and slightly higher forgetting (0.0340). This supports the subspace-preservation interpretation of ReCoLoRA: later tasks should not merely receive more low-rank capacity; they should be prevented from overwriting rank dimensions already used by previous tasks, and rank growth should not introduce unintended function drift.

\section{Ablations and Robustness Checks}\label{app:ablations}
The ablations isolate the design choices in ReCoLoRA. The Qwen3-8B ablations use randomized SVD initialization, elbow-based rank control, and a static two-stage principal-to-residual transition as the final baseline. \textbf{Dynamic recovery} adds the optional residual/rank recovery allocator during Stage 2. \textbf{One-stage} removes the principal-to-residual split and activates recovery from the beginning. \textbf{Fixed-rank} bypasses the elbow rank selector and forces a uniform rank of 16. \textbf{Random-init} replaces SVD initialization with random LoRA-style initialization while keeping the dynamic recovery setting. On Llama-3-8B, we further compare task-wise elbow rank expansion, layer/depth bonus expansion, and task-wise rank freezing. These variants test whether continual robustness comes merely from adding rank capacity or from preventing old task subspaces from being overwritten.
\subsection{Ablation Results}
\begin{figure}[t]
\centering
\includegraphics[width=0.7\linewidth]{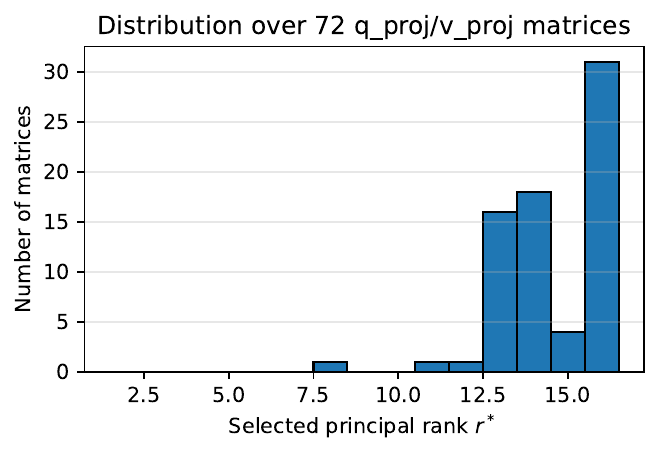}
\caption{Distribution of the selected principal rank $r^*$ across all 72 \texttt{q\_proj}/\texttt{v\_proj} matrices in Qwen3-8B under the elbow/energy criterion of Eq.~\ref{eq:elbow} ($\rho=0.8$, $r_{\max}=16$, seed 42). Every matrix selects $r^*\in[8,16]$ (mean $14.54$), confirming the $r\in[8,16]$ range used throughout the paper from the actual model weights rather than an assumed range.}
\label{fig:elbow-hist}
\end{figure}
Table~\ref{tab:qwen3-ablation} reports the completed ablations. The final ReCoLoRA configuration is the static two-stage variant, which uses SVD initialization, elbow/energy rank control ($\rho=0.8$ over the truncated spectral budget, $r\in[8,16]$ per projection matrix; Figure~\ref{fig:elbow-hist} shows the empirical distribution of $r^*$ across all targeted Qwen3-8B matrices), and the principal-to-residual stage transition. This variant achieves the best final average score and the best forgetting among all variants in the table, while also using fewer trainable parameters (6,850,688 on average) than the fixed-rank variant (7,668,864).

The strongest negative result is the one-stage variant. Removing the principal-to-residual schedule reduces the final average score from 0.8821 to 0.6548 and increases forgetting from 0.0240 to 0.2331. This strongly supports the central scheduling hypothesis: activating the residual/recovery mechanism too early makes training unstable in the continual setting.

SVD initialization remains useful: the random-init variant (0.8497/0.0455) is worse than ReCoLoRA (ours) (0.8821/0.0240) in both final average score and forgetting, even though both use the same elbow-selected per-layer ranks, which suggests that aligning the adapter with pretrained principal directions improves continual robustness independently of rank allocation. The fixed-rank comparison is informative for the first time: forcing a uniform rank of 16 at every layer (7,668,864 trainable parameters) yields both a lower final average score and higher forgetting (0.8658/0.0407) than the elbow-selected per-layer ranks of ReCoLoRA (ours) (6,850,688 parameters on average, $r\in[8,16]$), despite using more parameters. This supports the elbow/energy rank selector as a genuine accuracy and stability improvement rather than merely a parameter-reduction heuristic. The optional dynamic recovery allocator (0.8566/0.0466) remains weaker than static two-stage ReCoLoRA, so we continue to treat it as an extension that requires further tuning rather than a core component.

\begin{table}[t]
\centering
\begin{tabular}{lccc}
\toprule
Variant & Final Avg. $\uparrow$ & Avg. Forgetting $\downarrow$ & Trainable Params \\
\midrule
ReCoLoRA (ours) & \textbf{0.8821 $\pm$ 0.0058} & \textbf{0.0240 $\pm$ 0.0054} & 6,850,688$^{*}$ \\
Dynamic recovery & 0.8566 $\pm$ 0.0175 & 0.0466 $\pm$ 0.0190 & 6,850,688$^{*}$ \\
One-stage & 0.6548 $\pm$ 0.2405 & 0.2331 $\pm$ 0.2338 & 6,850,688$^{*}$ \\
Fixed rank & 0.8658 $\pm$ 0.0211 & 0.0407 $\pm$ 0.0224 & 7,668,864 \\
Random init & 0.8497 $\pm$ 0.0052 & 0.0455 $\pm$ 0.0108 & 6,850,688$^{*}$ \\
\bottomrule
\end{tabular}
\caption{Qwen3-8B ReCoLoRA ablations on the continual GLUE sequence. Results are mean $\pm$ standard deviation over seeds 42, 43, and 44. Static two-stage ReCoLoRA is the final method; the optional dynamic recovery allocator is weaker in this setting. $^{*}$Trainable parameters vary slightly by seed (6,848,640--6,851,712; average shown) because the elbow/energy rank selector assigns a different principal rank $r\in[8,16]$ per projection matrix (see Table~\ref{tab:qwen3-main}). Fixed rank forces $r=16$ uniformly and is therefore unaffected.}
\label{tab:qwen3-ablation}
\end{table}

\subsection{Robustness Check: Energy-Threshold Sensitivity}
\label{sec:robustness-checks}
The main results (Table~\ref{tab:qwen3-main}) use a single energy threshold $\rho=0.8$. We report a Qwen3-8B check, using the same training budget, hyperparameters, and seeds (42, 43, 44) as Table~\ref{tab:qwen3-main}, that probes the sensitivity of the main results to this choice.

\begin{figure}[t]
\centering
\includegraphics[width=\linewidth]{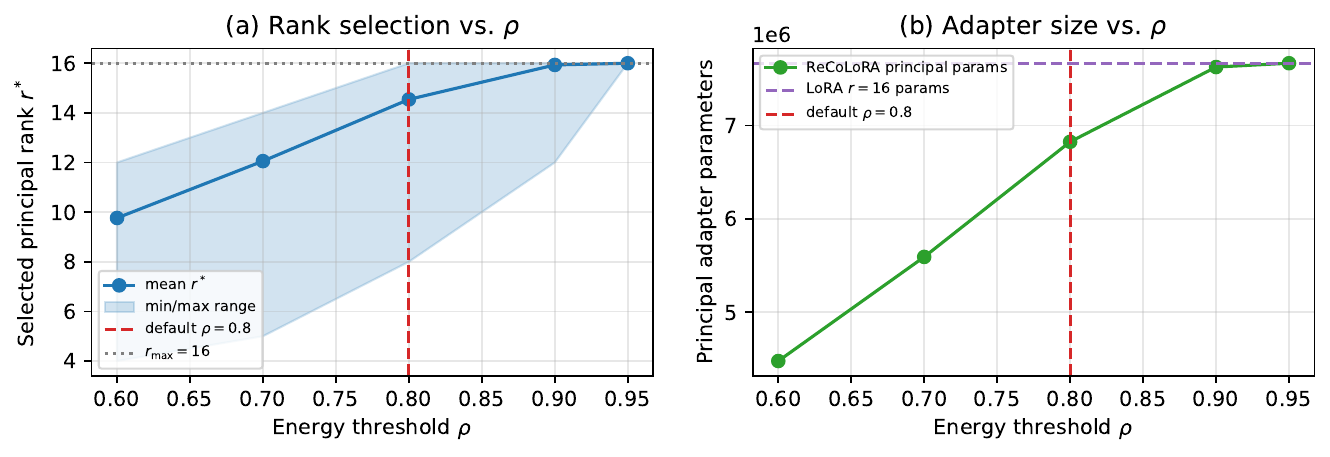}
\caption{Sensitivity of elbow/energy rank selection (Eq.~\ref{eq:elbow}) to the energy threshold $\rho$, computed from the same 72 Qwen3-8B \texttt{q\_proj}/\texttt{v\_proj} matrices and SVD (seed 42) as Figure~\ref{fig:elbow-hist}. (a) Mean selected rank $r^*$ (line) and per-matrix min/max range (shaded band) versus $\rho$; at $\rho=0.95$ every matrix saturates at $r_{\max}=16$. (b) Resulting principal-adapter parameter count versus $\rho$, with LoRA's $r=16$ parameter count shown for reference.}
\label{fig:rho-sensitivity}
\end{figure}

\paragraph{Energy-threshold sensitivity.} Figure~\ref{fig:rho-sensitivity} sweeps the energy threshold $\rho\in\{0.6,0.7,0.8,0.9,0.95\}$ in Eq.~\ref{eq:elbow}, holding the randomized SVD fixed (same 72 \texttt{q\_proj}/\texttt{v\_proj} matrices and seed as Figure~\ref{fig:elbow-hist}) and recomputing $r^*$ for each $\rho$. As $\rho\to0.95$, every matrix saturates at $r^*=r_{\max}=16$, so ReCoLoRA's principal adapter becomes parameter-equivalent to the Fixed-rank ablation in Table~\ref{tab:qwen3-ablation}; as $\rho\to0.6$, the mean selected rank drops to $9.76$ (range $[4,12]$), roughly $65\%$ of the $\rho=0.8$ adapter size. Table~\ref{tab:qwen3-rho-sensitivity} reports downstream continual fine-tuning at $\rho\in\{0.6,0.9\}$ on the main task order. Final average score and forgetting both improve monotonically as $\rho$ increases over $\{0.6,0.8,0.9\}$, and $\rho=0.9$ slightly outperforms the default $\rho=0.8$ on both metrics, though the two are within one standard deviation of each other. Notably, even at $\rho=0.9$ (mean $r^*=15.93$, $7{,}627{,}904$ trainable parameters, within $1\%$ of LoRA's $7{,}667{,}712$), ReCoLoRA still outperforms LoRA by a wide margin on both axes ($0.8849$ vs.\ $0.8634$ final average, $0.0197$ vs.\ $0.0414$ forgetting). This suggests that ReCoLoRA's gain over LoRA comes mainly from \emph{per-matrix adaptive rank and spectral initialization} rather than from aggressively shrinking the adapter, and that $\rho=0.8$ is a reasonable but not provably optimal default; within this range, $\rho=0.9$ is a plausible alternative when forgetting is the priority.

\begin{table}[t]
\centering
\begin{tabular}{lcccc}
\toprule
$\rho$ & mean $r^*$ & Final Avg. $\uparrow$ & Avg. Forgetting $\downarrow$ & Trainable Params \\
\midrule
0.6 & 9.76 & 0.8750 $\pm$ 0.0059 & 0.0324 $\pm$ 0.0083 & 4,472,960 \\
0.8 (default) & 14.54 & 0.8821 $\pm$ 0.0058 & 0.0240 $\pm$ 0.0054 & 6,850,688$^{*}$ \\
0.9 & 15.93 & \textbf{0.8849 $\pm$ 0.0036} & \textbf{0.0197 $\pm$ 0.0025} & 7,627,904 \\
\bottomrule
\end{tabular}
\caption{ReCoLoRA energy-threshold $\rho$ sensitivity on Qwen3-8B, seeds 42, 43, 44 (the $\rho=0.8$ row reproduces Table~\ref{tab:qwen3-main}). Mean $r^*$ is averaged over the 72 \texttt{q\_proj}/\texttt{v\_proj} matrices used to build Figure~\ref{fig:rho-sensitivity}. $^{*}$Trainable parameters vary slightly by seed (6,848,640--6,851,712; average shown).}
\label{tab:qwen3-rho-sensitivity}
\end{table}

\bibliographystyle{unsrt}
\bibliography{references}

\begin{thebibliography}{10}

\bibitem{hu2022lora}
Edward~J. Hu, Yelong Shen, Phillip Wallis, Zeyuan Allen-Zhu, Yuanzhi Li, Shean
  Wang, Lu~Wang, and Weizhu Chen.
\newblock Lora: Low-rank adaptation of large language models.
\newblock In {\em Proc. ICLR}, 2022.

\bibitem{wang2024comprehensivecl}
Liyuan Wang, Xingxing Zhang, Hang Su, and Jun Zhu.
\newblock A comprehensive survey of continual learning: Theory, method and
  application.
\newblock {\em arXiv preprint arXiv:2302.00487}, 2024.

\bibitem{ke2023continualnlp}
Zixuan Ke and Bing Liu.
\newblock Continual learning of natural language processing tasks: A survey.
\newblock {\em arXiv preprint arXiv:2211.12701}, 2023.

\bibitem{luo2023empiricalforgetting}
Yun Luo, Zhen Yang, Fandong Meng, Yafu Li, Jie Zhou, and Yue Zhang.
\newblock An empirical study of catastrophic forgetting in large language
  models during continual fine-tuning.
\newblock {\em arXiv preprint arXiv:2308.08747}, 2023.

\bibitem{weng2024continualllm}
Tongtong Wu, Linhao Luo, Yuan-Fang Li, Shirui Pan, Thuy-Trang Vu, and
  Gholamreza Haffari.
\newblock Continual learning for large language models: A survey.
\newblock {\em arXiv preprint arXiv:2402.01364}, 2024.

\bibitem{zhang2023adalora}
Qingru Zhang, Minshuo Chen, Alexander Bukharin, Pengcheng He, Yu~Cheng, Weizhu
  Chen, and Tuo Zhao.
\newblock Adaptive budget allocation for parameter-efficient fine-tuning
  (adalora).
\newblock In {\em Proc. ICLR}, 2023.

\bibitem{meng2024pissa}
Fanxu Meng, Zhaohui Wang, and Muhan Zhang.
\newblock Pissa: Principal singular values and singular vectors adaptation of
  large language models.
\newblock {\em arXiv preprint arXiv:2404.02948}, 2024.

\bibitem{liu2024dora}
Shih-Yang Liu, Chien-Yi Wang, Hongxu Yin, Pavlo Molchanov, Frank Wang,
  Kwang-Ting Cheng, and Min-Hung Chen.
\newblock Dora: Weight-decomposed low-rank adaptation.
\newblock In {\em Proc. ICML}, 2024.

\bibitem{wang2023olora}
Xiao Wang, Tianze Chen, Qiming Ge, Han Xia, Rong Bao, Rui Zheng, Qi~Zhang, Tao
  Gui, and Xuanjing Huang.
\newblock Orthogonal subspace learning for language model continual learning.
\newblock In {\em Findings of EMNLP}, 2023.

\bibitem{wang2019glue}
Alex Wang, Amanpreet Singh, Julian Michael, Felix Hill, Omer Levy, and
  Samuel~R. Bowman.
\newblock Glue: A multi-task benchmark and analysis platform for nlu.
\newblock In {\em EMNLP Workshop BlackboxNLP}, 2019.

\bibitem{wang2019superglue}
Alex Wang, Yada Pruksachatkun, Nikita Nangia, Amanpreet Singh, Julian Michael,
  Felix Hill, Omer Levy, and Samuel~R. Bowman.
\newblock Superglue: A stickier benchmark for general-purpose language
  understanding systems.
\newblock In {\em NeurIPS}, 2019.

\bibitem{sun2019lamol}
Fan-Keng Sun, Cheng-Hao Ho, and Hung-Yi Lee.
\newblock {LAMOL}: {LAnguage MOdeling} for lifelong language learning.
\newblock {\em arXiv preprint arXiv:1909.03329}, 2019.

\bibitem{scialom2022continualt0}
Thomas Scialom, Tuhin Chakrabarty, and Smaranda Muresan.
\newblock Fine-tuned language models are continual learners.
\newblock In {\em Proc. EMNLP}, 2022.

\bibitem{valipour2023dylora}
Mojtaba Valipour, Mehdi Rezagholizadeh, Ivan Kobyzev, and Ali Ghodsi.
\newblock Dylora: Parameter-efficient tuning of pre-trained models using
  dynamic search-free low-rank adaptation.
\newblock In {\em Proc. EACL}, 2023.

\bibitem{kopiczko2024vera}
Dawid~J. Kopiczko, Tijmen Blankevoort, and Yuki~M. Asano.
\newblock Vera: Vector-based random matrix adaptation.
\newblock In {\em Proc. ICLR}, 2024.

\bibitem{hayou2024loraplus}
Soufiane Hayou, Nikhil Ghosh, and Bin Yu.
\newblock Lora+: Efficient low rank adaptation of large models.
\newblock {\em arXiv preprint arXiv:2402.12354}, 2024.

\bibitem{razdaibiedina2023progressiveprompts}
Anastasia Razdaibiedina, Yuning Mao, Rui Hou, Madian Khabsa, Mike Lewis, and
  Amjad Almahairi.
\newblock Progressive prompts: Continual learning for language models.
\newblock In {\em Proc. ICLR}, 2023.

\bibitem{zhang2025clora}
Xin Zhang, Liang Bai, Xian Yang, and Jiye Liang.
\newblock {C-LoRA}: Continual low-rank adaptation for pre-trained models.
\newblock {\em arXiv preprint arXiv:2502.17920}, 2025.

\bibitem{halko2011finding}
Nathan Halko, Per-Gunnar Martinsson, and Joel~A. Tropp.
\newblock Finding structure with randomness: Probabilistic algorithms for
  constructing approximate matrix decompositions.
\newblock {\em SIAM Review}, 53(2):217--288, 2011.

\end{thebibliography}
\end{document}